\documentclass[conference]{IEEEtran}
\IEEEoverridecommandlockouts

\usepackage{cite}
\usepackage{amsmath,amssymb,amsfonts}
\usepackage{algorithmic}
\usepackage{graphicx}
\usepackage{textcomp}
\usepackage{xcolor}
\usepackage{booktabs}  
\def\BibTeX{{\rm B\kern-.05em{\sc i\kern-.025em b}\kern-.08em
    T\kern-.1667em\lower.7ex\hbox{E}\kern-.125emX}}
\begin{document}

\title{Unmasking Facial DeepFakes: A Robust Multiview Detection Framework for Natural Images \\
}

\author{\IEEEauthorblockN{1\textsuperscript{st} Sami Belguesmia }
\IEEEauthorblockA{\textit{Dep. of Computer science and Eng.} \\
\textit{University of Quebec in Outaouais}\\
Gatineau, Canada \\
 bels106@uqo.ca}
\and
\IEEEauthorblockN{2\textsuperscript{nd} Mohand Said Allili}
\IEEEauthorblockA{\textit{Dep. of Computer science and Eng.} \\
\textit{University of Quebec in Outaouais}\\
Gatineau, Canada \\
 mohandsaid.allili@uqo.ca}
\and
\IEEEauthorblockN{3\textsuperscript{rd} Assia Hamadene}
\IEEEauthorblockA{\textit{Dep. of Computer science and Eng.} \\
\textit{University of Quebec in Outaouais}\\
Gatineau, Canada \\
 assia.hamadene@uqo.ca}
}


\maketitle

\begin{abstract}
DeepFake technology has advanced significantly in recent years, enabling the creation of highly realistic synthetic face images. Existing DeepFake detection methods often struggle with pose variations, occlusions, and artifacts that are difficult to detect in real-world conditions. To address these challenges, we propose a multi-view architecture that enhances DeepFake detection by analyzing facial features at multiple levels. Our approach integrates three specialized encoders, a global view encoder for detecting boundary inconsistencies, a middle view encoder for analyzing texture and color alignment, and a local view encoder for capturing distortions in expressive facial regions such as the eyes, nose, and mouth, where DeepFake artifacts frequently occur. Additionally, we incorporate a face orientation encoder, trained to classify face poses, ensuring robust detection across various viewing angles. By fusing features from these encoders, our model achieves superior performance in detecting manipulated images, even under challenging pose and lighting conditions. Experimental results on challenging datasets demonstrate the effectiveness of our method, outperforming conventional single-view approaches.
\end{abstract}


\section{Introduction}

In recent years, the rise of DeepFake technology has led to the generation of highly realistic synthetic images, posing significant challenges to security, digital forensics, and media authenticity \cite{Wang2024}. DeepFake images, created using advanced generative models such as Generative Adversarial Networks (GANs) \cite{Karras2019} and Variational Autoencoders (VAEs) \cite{Razavi2019}, can seamlessly alter or fabricate human faces, making them nearly indistinguishable from real photographs. Recently, diffusion models \cite{Blattmann2023} have greatly enhanced the generation capability of images and videos. While these synthetic images have applications in entertainment and creative industries, the proliferation of deepfake technology poses escalating risks to biometric security, online misinformation, identity fraud, and manipulation of public perception. 

DeepFakes can generally be classified into four mainstream categories \cite{Pei2024}, each leveraging advanced deep learning techniques to manipulate facial images and videos: (1) Face Swapping \cite{Ancilotto2023,Shiohara2023}, which replaces a person's face with that of another individual while preserving the background and overall scene consistency; (2) Face Reenactment \cite{Zhang2023}, which transfers facial expressions, emotions, and head movements from a reference video onto a static facial image while maintaining the original identity of the source; (3) Talking Face Generation \cite{Sheng2023}, which synchronizes mouth movements with a given speech or textual input to create highly realistic, lip-synced animations; and (4) Facial Attribute Editing \cite{Huang2023,Ning2023}, which modifies specific facial features such as age, gender, hairstyle, or skin tone while preserving the subject’s overall identity. Over time, DeepFake generation models have significantly improved in quality, particularly with the introduction of diffusion models \cite{Blattmann2023}, which enhanced realism and reduced detectable artifacts. Additionally, the scope of synthetic content has expanded beyond single-frame image generation to temporal video synthesis, enabling more seamless and dynamic manipulations that pose increasing challenges for detection and forensic analysis.

Detecting DeepFake face images has become a critical research area, with numerous approaches proposed to address this challenge \cite{Wang2024}. Traditional methods rely on handcrafted features, such as inconsistencies in facial texture, lighting, or eye reflections \cite{Yan2024,Allili2012,Yapi2023}. These methods  typically analyze entire faces using single-stream feature extraction, but modern deepfake techniques reduced  global inconsistencies, which makes difficult to describe Deepfakes holistically. Recent learning-based techniques, particularly Convolutional Neural Networks (CNNs) and Transformer-based models, have demonstrated superior performance in identifying DeepFake anomalies \cite{Pei2024}. These models leverage spatial, temporal, and frequency-domain features to detect subtle artifacts introduced during the synthesis process \cite{Frank2020}. Additionally, hybrid approaches that integrate multiple views, including facial orientation, biological signals, and multimodal feature fusion, have shown promise in enhancing detection accuracy \cite{Sun2024}. 

Despite significant progress, DeepFake detection remains an evolving field, as generative models continue to improve, reducing detectable artifacts \cite{Tan2024,Allili2019}. Most DeepFake detection methods assume that the face in an image is un-oriented, meaning that the subject's viewing direction is primarily facing the camera \cite{Wang2024}. This simplifies feature extraction, as many deep learning models are trained on frontal-facing images. However, in real-world scenarios, face orientations can vary significantly due to uncontrolled acquisition conditions such as pose variation, occlusion, illumination changes, and motion blur, making detection more challenging \cite{Le2021}. Faces may appear at different angles, including side profiles or extreme head tilts, which can obscure subtle DeepFake artifacts. Additionally, occlusions caused by objects like hair, glasses, or hands can lead to incomplete facial information, while inconsistent lighting and shadowing effects may either mask or exaggerate DeepFake anomalies. To overcome these challenges, researchers are exploring methods relying on pose-invariant DeepFake detection strategies, or extracting  head pose estimation patterns to enhance the robustness of deepfake detection \cite{Becattini2024}. Techniques like facial landmark detection and 3D Morphable Models (3DMMs) \cite{Li2024} enable face alignment and normalization before applying detection algorithms.

In this work, we propose a multi-view framework that enhances the robustness of DeepFake face detection in the wild by integrating multi-view feature extraction and face orientation analysis. Our key contribution relies on the hypothesis that most DeepFake manipulations—such as face swapping and face reenactment—are primarily localized in the central facial region, where artifacts and inconsistencies are often present. By leveraging multiple views of the face, our approach improves the detection of subtle manipulations that might be overlooked in single-view analysis. Specifically, our method systematically examines both global and local facial features, capturing inconsistencies across different spatial scales. To validate this hypothesis, we designed and compared two variants of our multi-view model: one leveraging Convolutional Neural Networks (CNNs) for detailed spatial feature extraction and the other utilizing vision Transformers for improved global contextual understanding. We conducted extensive evaluations on two challenging benchmark datasets, OpenForensics \cite{Le2021} and FaceForensics++ \cite{rossler2019faceforensics}, demonstrating promising performance and highlighting the effectiveness of our approach in detecting DeepFakes across diverse real-world scenarios.

The rest of the paper is organized as follows: Section \ref{review} presents some related work. Section \ref{methodology} presents the proposed methodology. We end the paper with a conclusion and future work perspectives. 

\section{Related work}
\label{review}
To deal with the increasingly realistic manipulated faces threat, various researches have been carried out with a large effort to provide efficient detection systems \cite{Kumar2025}. Several comprehensive surveys review the deepfakes detection and generation literature through several taxonomies and  categorization from different perspectives \cite{TOLOSANA2020} \cite{Naitali2023} \cite{Zahid2023} \cite{Abbas2024} \cite{Nguyen2022}. For instance, Verdoliva in \cite{Verdoliva2020} examined detection methods through calssical approaches such as supervised methods with handcrafted features and deep learning-based approaches such as CNN models. 
The authors in \cite{TOLOSANA2020} reviewed face manipulation techniques and fake detection system and categorized both creation and detection methods based on deepfakes creation manner such as identity swap, attribute manipulation, expression swap and entirely synthesized faces.  The authors in \cite{Mirsky2021} exclusively focused on deepfakes pertaining to the human face and body such as changing a target’s expression or a body or face part such as eyes, gaze, as well as face replacement by swapping or transfer methods. Whereas, Abbas  et al in \cite{Abbas2024}  analysed detection methods for image, audio, and video detection. They grouped deepfake detection into two sections, the face swap and face reenactment deepfake detection on one hand and  the synthetic faces and audio-visual detection on the other hand. They also categorized each section according to two subcategories; the deep learning-based approaches and machine learning-based approaches. 
A different perspective and taxonomy is carried out by Nguyen et al in  \cite{Nguyen2022} where the authors present an overview focusing on whether the type of data is images or videos. The fake images detection is revised through handcrafted features and deep features separately whereas for fake video detection, they present two main subcategories based on whether the temporal characteristics or the visual ones are exploited within the video frame.

More recently, a comprehensive survey offered an overview with a particular focus on  ViT-based deepfake detection models as well as a concise description of the structure and key characteristics of each model \cite{Wang2024ATS}.  For example, one of the relevant ViT deepfakes systems is the Identity Consistency Transformer (ICT) proposed by Dong et al in \cite{Dong2022}, the system  relies on inner and outer face regions to detect identities inconsistency and achieves high performances across several datasets, as well as across several manipulation types, a reference-assisted version of the proposed method further improves detection efficiency.   However,  the ICT method may not be efficient when confronted to consistent identities \cite{Wang2024ATS}. The authors in \cite{Wang2022} proposed a Multi-modal Multi-scale Transformer (M2TR), where the system uses patches with different sizes and multi-scale levels to capture image artifacts. This method underlines the crucial role of  the ViTs in enhancing deepfakes detection systems. Finally, hybrid models such as combining Convolutional Neural Networks and sequential models allows to take advantage of each model’s strength by extracting low-level features and long dependencies \cite{Jayashre2024}. 

\section{Methodology}
\label{methodology}

\subsection{Image pre-processing}
Our methodology begins with an input image being processed by the RetinaFace detector \cite{Deng2020}, which  identifies all faces within the image and extracts five key facial landmarks for each face : two for the eyes, one for the nose, and two defining the boundaries of the mouth. These landmarks serve as critical reference points for precise facial alignment and normalization. The bounding box of the face serve as our middle view.
\begin{figure}[hb]
    \centering
    \includegraphics[width=0.90\linewidth]{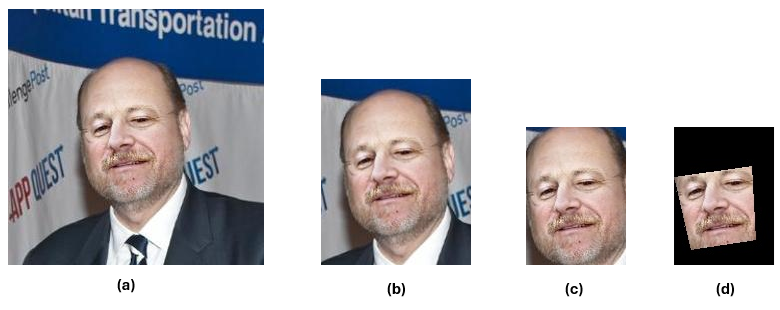}
    \caption{Illustration of the image pre-processing creating : (b) global, (c) middle, and  (d) local views, respectively, from the original image (a).}
    \label{preprocessing}
\end{figure}
To obtain local face view, we  construct a convex hull around the five detected landmarks, expanding the region by a 15-pixel margin to fully encompass the eyes and mouth while preserving essential facial details (see Fig. \ref{preprocessing} for illustration). This localized crop helps focus on fine-grained features, such as eyes, noise and mouth alignment and blending artifacts, that are crucial for DeepFake detection. Then, the global view is obtained by expanding the middle view in each direction by  20 pixels. This view will contain the neck and the ears and some of the background region.

Finally, both the global and local face images are resized to $224 \times 224$ pixels, with zero-padding applied to maintain the aspect ratio and preserve original content. This pre-processing step ensures that the extracted facial features remain consistent and optimally structured for downstream analysis by the multi-view encoders.

\subsection{The general multi-view architecture}
Our proposed architecture is summarized in Fig. \ref{fig:architecture}, which leverages three encoders extracting face information at different levels to identify DeepFakes. To identity deepfake artifacts, the face is processed  through a multi-view embedding process, consisting of three specialized sub-encoders, each designed to capture complementary facial attributes, in addition to a fourth branch encoding the pose of the face. Our proposed multi-view encoding architecture is designed to capture DeepFake artifacts at different levels of facial representation, enhancing detection robustness against diverse manipulation techniques: 

\begin{figure*}[htb]
    \centering
    \includegraphics[width=0.98\textwidth]{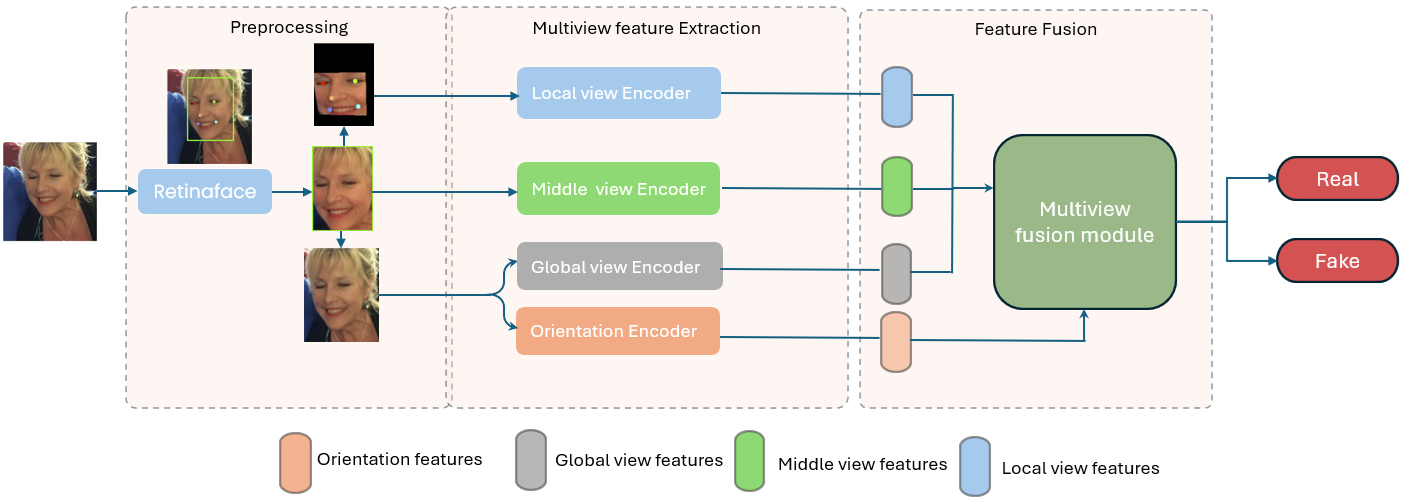}
    \caption{Overview of the proposed three stream architecture. The global and middle  stream extract holistic facial representations, while the local region-specific stream focuses on high-importance facial sub-regions.}
    \label{fig:architecture}
\end{figure*}

\begin{figure}[htb]
I    \centering
    \includegraphics[width=0.95\linewidth]{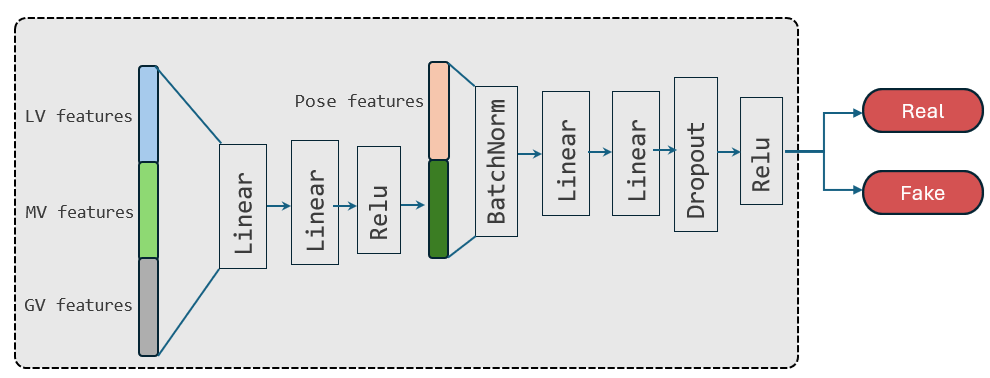}
    \caption{Illustration of fusion module integrating local view (LV), middle view (MV), global view (GV) and pose features.}
    \label{Fusion_module}
\end{figure}

\begin{itemize}
    \item \textit{The global view encoder}: analyzes the entire facial structure, including the face boundaries, which often exhibit inconsistencies in cases of face swapping or facial attribute editing. This encoder helps identify unnatural transitions between the face and the surrounding context, such as blending inconsistencies, edge distortions, or unnatural lighting variations.  

    \item \textit{The middle view encoder:} focuses on facial texture, color consistency, and part alignment, extracting fine-grained details to detect subtle skin tone mismatches, lighting irregularities, and blending artifacts introduced by DeepFake synthesis. This level of analysis ensures that manipulated areas, such as smoothed-over skin regions or artificial lighting corrections, do not go unnoticed.  

    \item  \textit{The local view encoder:} is dedicated to detecting artifacts in the most expressive and manipulation-prone regions of the face, specifically the eyes, nose, and mouth. These areas are particularly vulnerable to distortions from face reenactment and talking face generation, where artifacts such as geometric asymmetries, unnatural mouth movements, inconsistent blinking, or misaligned eye gaze often occur. By focusing on these critical regions, the local encoder enhances the model's ability to spot unnatural expressions and facial dynamics. 
   
    \item  \textit{Pose encoder}:  Additionally, a fourth encoder is trained to estimate the face orientation, classifying it into one of 13 different poses, covering 10 tilt and 10 pan angles. This orientation encoder enables the model to handle variations in face alignment, ensuring that DeepFake artifacts remain detectable even when the subject is not facing the camera directly.  
\end{itemize}

Fig. \ref{Fusion_module} illustrates the fusion module, which integrates feature representations extracted from the three view encoders and the orientation encoder to enhance DeepFake detection. This module is designed as a multi-layer perceptron (MLP) and consists of two main parts. The first parts focuses on merging the features extracted from the three view encoders using a sequence of two linear layers followed by a non-linear activation function, effectively capturing cross-view relationships. The second component incorporates both face content and pose features through a structured pipeline that begins with batch normalization to stabilize feature distributions, followed by two fully connected layers interleaved with dropout and non-linear activation functions to enhance generalization. This fusion strategy ensures a comprehensive and discriminative feature representation, enabling the model to robustly detect deepfake artifacts across varying facial orientations. 

\subsection{Model initialization and training}

The proposed model is initialized using pretrained weights leveraging fine-tuning for improved feature extraction. For the CNN-based variant, we used  a pre-trained ResNet50 model \cite{He2016} fort each view encoder, where we retain the feature extraction layers and introduce a non-linear classification head for each encoder. For Transformers-based variant, we used a pre-trained BeiT  model \cite{Bao2022} for each view encoder. The encoders are designed to extract multi-scale facial features from different views. Thus, each encoder is trained using its corresponding  local/middle/global face images, capturing global to fine-grained details. The local view encoder analyses subtle DeepFake traces in  localized facial regions. The middle and global view encoders aim at analyzing global face characteristics such as texture inconsistencies, blending artifacts, ensuring that holistic facial structures, overall symmetry, and contextual relationships within the face are authentic. The fourth encoder is a MobileNet, pre-trained on a separate dataset to predict 13 different facial poses, allowing the model to account for pose variations and orientation mismatches that could otherwise hinder DeepFake detection. All models were trained using the binary cross entropy loss and the Adam Optimizer, for 100 epochs and a learning rate of $0.0001$.

\section{Experimental results}
To validate the proposed framework,  we performed several experiments using recent popular benchmarks. We also provide quantitative results comparing our framework with existing work. In what follows, we describe the used datasets and the quantitative metics used for evaluation. 

\subsection{Datasets}
To validate our method, we used first the OpenForensics \cite{Le2021} and FaceForensics++ \cite{rossler2019faceforensics} datasets, which is specifically designed for realistic and complex scenarios. The OpenForensics dataset contains a diverse set of images featuring faces captured in uncontrolled environments, distinguishing it from previous datasets that often lacked diversity or realism. Additionally, each image may contain one or multiple faces, making it well-suited for evaluating DeepFake detection models under real-world conditions. The FaceForensics dataset  contains approximately  400 videos with varying length, where each video is captured for a moving subject against a background. For this dataset, we selected one frame per 10 in each video, and merged all frames to constitute our dataset. For each image, we applied facial detection and landmark extraction to identify key facial regions, facilitating the pre-processing required for multi-view feature extraction in our DeepFake detection framework. To evaluate the performance of the models, we used the precision, recall, F1 score score metrics as well as Area under curve (AUC), where we split each dataset using $70\%$ for training, $15\%$ for validation and $15\%$ for testing.   

\begin{figure}
    \centering
    \includegraphics[width=0.95\linewidth]{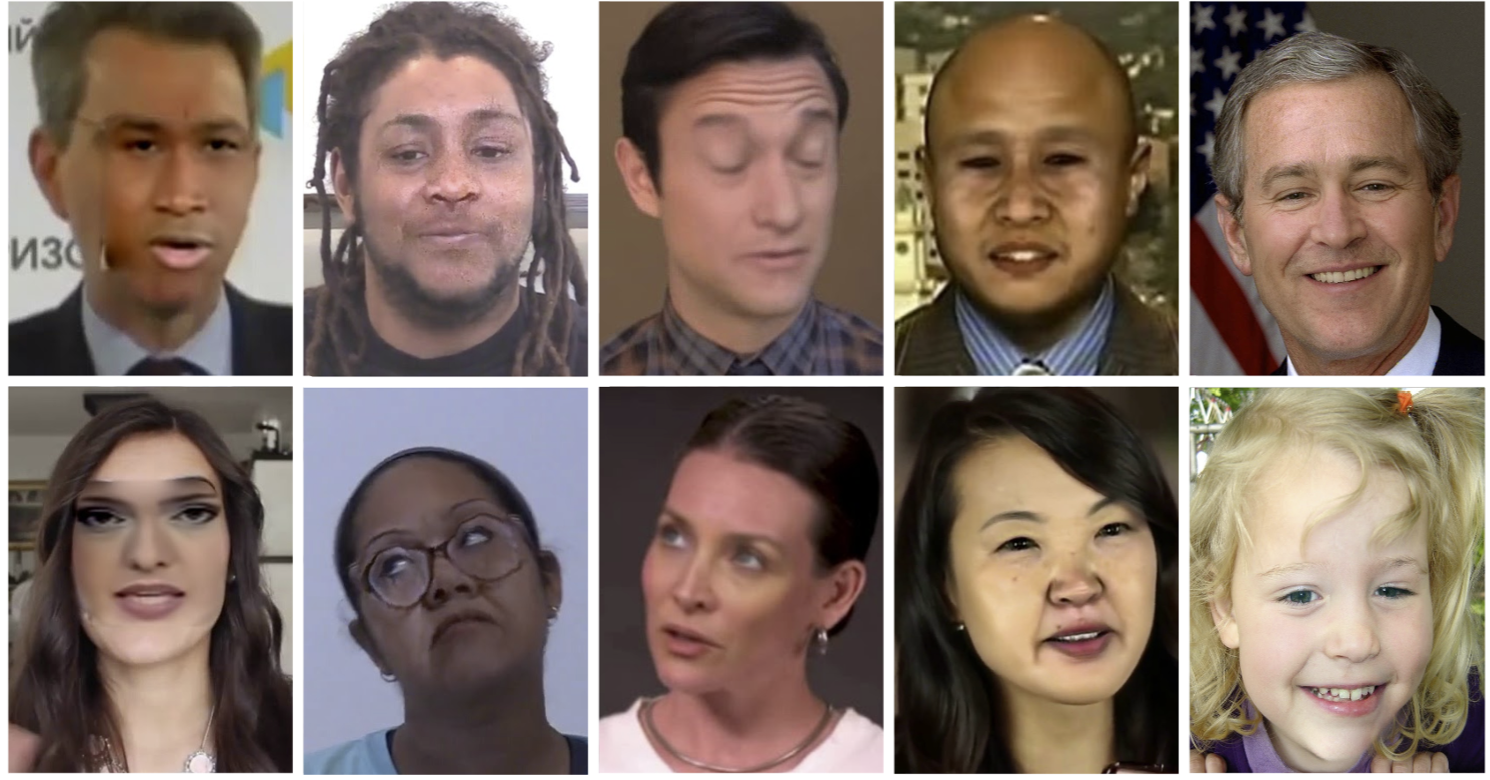}
    \caption{Samples from different DeepFake datasets. From left to right, FaceForensics++ \cite{rossler2019faceforensics}, DFDC \cite{dolhansky2020deepfake}, DeeperForensics \cite{jiang2020deeperforensics}, Celeb-DF \cite{li2020celebdf} and OpenForensics \cite{Le2021}.}
    \label{fig:dataset_comparison}
\end{figure}

\subsection{Quantitative results}

To thoroughly assess the effectiveness of our approach, we conducted an ablation study comparing different views and Deepfake classification without incorporating orientation information. Tables \ref{tab:comparison1} and \ref{tab:comparison2} present the comparative results obtained using the OpenForensics \cite{Le2021} and FaceForensics++ \cite{rossler2019faceforensics} datasets, respectively. The results clearly demonstrate that, for both datasets, the fusion of multiple views significantly enhances performance compared to any single-view implementation. This highlights the advantage of incorporating diverse perspectives, which allows the model to capture richer feature representations and mitigate biases introduced by a single viewpoint.

Furthermore, the comparative analysis between CNN-based and BeiT-based architectures reveals interesting insights. While both architectures benefited from multi-view integration, the CNN-based variant exhibited slightly better performance than the BeiT-based counterpart in our experiments. This suggests that CNNs, which are traditionally optimized for spatial feature extraction, may still hold an edge in scenarios where local texture and fine-grained structural information play a dominant role in Deepfake detection. However, the BeiT-based model demonstrated competitive performance, indicating its potential to leverage self-attention mechanisms for improved contextual understanding.

Additionally, our findings underscore the importance of orientation-aware feature extraction, as excluding orientation information led to a noticeable drop in classification performance across both datasets. This reinforces the hypothesis that considering spatial and directional cues contributes to a more robust and discriminative feature representation. Future work could further explore hybrid architectures that combine CNNs strength in spatial representation with transformers ability to model long-range dependencies, potentially achieving an optimal balance between local and global feature extraction.

Finally, we compared our full models against two recent state-of-the-art methods evaluated on the same datasets, Lin et al.~\cite{Lin2024} and Concas et al.~\cite{Concas2024}. Our AUC scores show a slight but consistent advantage, confirming the effectiveness of integrating multiple views with orientation-aware features to learn more discriminative representations. The gains persist across datasets, underscoring robustness and stronger generalization than existing techniques. Looking ahead, incorporating contrastive learning (e.g., supervised or multi-view contrastive pretraining on content–pose pairs) \cite{Valem2024} can further tighten inter-class separation, mitigate domain shift, and amplify these improvements. Taken together, our approach offers a strong baseline for future Deepfake detection, especially when fine-grained cues and multi-perspective analysis are critical.

\begin{table}[h]
    \centering
    \begin{tabular}{|l|c|c|c|c|}
    \hline
        \toprule
        \textbf{Method} & \textbf{Precision} & \textbf{Recall} & \textbf{F1-score } & AUC\\
        \midrule
    \hline
        Local view (CNN) & 97.58 \% & 97.58 \% & 97.58 \% & -\\
    \hline        
        Middle view (CNN)   & 97.35 \% & 97.34 \% & 97.34 \%  & -\\
    \hline        
        Global view (CNN)   & 97.40 \% & 98.04  \% & 98.02  \%  & -\\
    \hline
        View fusion (CNN)   & 98.27 \% & 98.27 \% & 98.27 \%  & -\\
    \hline
        Fusion + pose  (CNN) & 98.86 \% & 98.10 \% & 98.59 \%  & 98.49 \%\\
    \hline
        Fusion + pose  (BeiT) & 98.47 \% & 98.50 \% & 98.49 \% &  98.02 \% \\
    \hline 
      Lin et al. \cite{Lin2024} & -  &-  & -  & 98.01 \%\\
    \hline 

        \bottomrule
    \end{tabular}
    \caption{Comparative results using OpenForensics dataset \cite{Le2021}.}
    \label{tab:comparison1}
\end{table}

\begin{table}[h]
    \centering
    \begin{tabular}{|l|c|c|c|c|}
    \hline
        \toprule
        \textbf{Method} & \textbf{Precision} & \textbf{Recall} & \textbf{F1-score } & AUC\\
        \midrule
    \hline
        Local view (CNN) & 95.38 \% & 97.58 \% & 95.38 \% & -\\
    \hline        
        Middle view (CNN)   & 97.15 \% & 97.22 \% & 97.68 \%  & -\\
    \hline        
        Global view (CNN)   & 97.30 \% & 97.90  \% & 98.05  \%  & -\\
    \hline
        View fusion (CNN)   & 98.25 \% & 98.17 \% & 98.20 \%  & -\\
    \hline
        Fusion + pose  (CNN) & 98.55 \% & 98.95 \% & 98.78 \%  & 99.88\%\\
    \hline
        Fusion + pose  (BeiT) & 97.33 \% &  97.29 \% &  97.30  \% &  99.68 \% \\
    \hline 
     Concas et al. \cite{Concas2024} & -  &-  & -  & 99.80  \%\\
    \hline

        \bottomrule
    \end{tabular}
    \caption{Comparative results using FaceForensics++ dataset \cite{rossler2019faceforensics}.}
    \label{tab:comparison2}
\end{table}

\subsection{Qualitative results}
\begin{figure}
    \centering
    \includegraphics[width=0.95\linewidth]{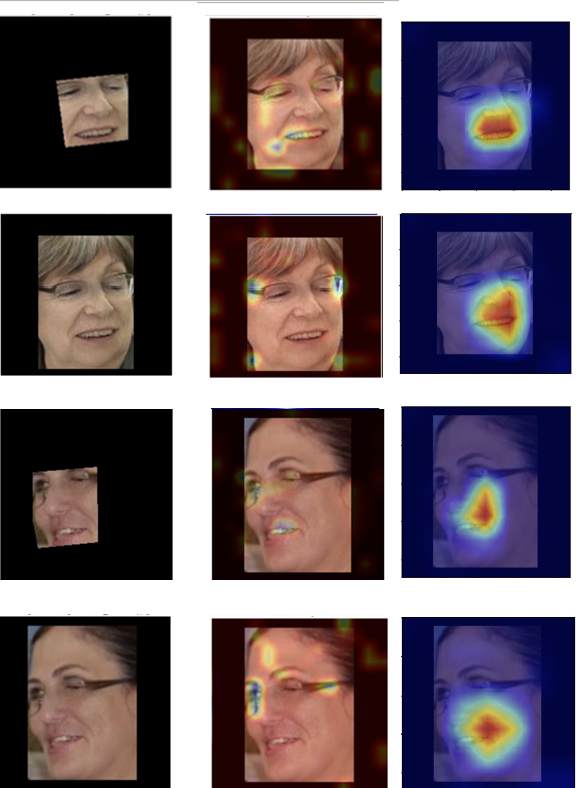}
    \caption{Examples of GradCam calculation  for two images using the BeiT (middle column) and CNN (right column) implementations.}
    \label{fig:enter-label}
\end{figure}

Fig. \ref{fig:enter-label} presents qualitative results using the Grad-CAM visualization, illustrating how different models focus on distinct image regions. While CNN-based models achieve high global accuracy, their attention maps tend to be more dispersed, sometimes failing to highlight critical fine-grained features. In contrast, the BeiT implementation demonstrates a more precise localization of detailed artifacts, indicating a stronger capability in capturing subtle patterns. This suggests that BeiT not only excels at recognizing fine-grained features but also provides more interpretable and explainable results, which is crucial for  DeepFake forensics, for example.

\section{Conclusion}
In this work, we proposed a multi-view encoder-based architecture for DeepFake detection, leveraging global, middle, and local facial feature extraction alongside a dedicated face orientation encoder to enhance robustness against adversarial manipulations. By integrating multi-scale feature analysis, our model effectively captures artifacts introduced by face manipulations, which are often difficult to detect using conventional methods. By fusing information from multiple views, our approach significantly improves generalization and robustness, making it effective for detecting DeepFakes in uncontrolled real-world environments, where factors such as pose variations, occlusions, and lighting inconsistencies present significant challenges. Experimental validation on challenging  datasets demonstrates the efficacy of our method in handling diverse facial representations, setting it apart from previous approaches that often struggle with pose and texture variability.  Future research can further enhance this approach by integrating temporal analysis for video-based detection, adversarial training for improved robustness, and explainable AI (XAI) techniques to increase interpretability. 


\section{Acknowledgment}
This work has been achieved thanks to the support of the Centre interdisciplinaire de recherche et d’innovation en cybersécurité et société (CIRICS) through the Fonds de recherche du Québec.

\end{document}